%% file: 0_main.tex
\documentclass{llncs}
\usepackage{makeidx}  
\usepackage{color}
\usepackage{multirow}
\usepackage{booktabs}
\usepackage{amsmath}
\usepackage{graphicx}
\usepackage{subcaption}
\DeclareMathOperator*{\argmax}{arg\,max}
\usepackage[pagebackref=true,breaklinks=true,letterpaper=true,colorlinks,bookmarks=false]{hyperref}

\begin{document}

\mainmatter              
%

\title{Food Ingredients Recognition \\through Multi-label Learning}
\titlerunning{Hamiltonian Mechanics}  
%
\author{Marc Bola\~nos\inst{1,2} \and Aina Ferr\`a\inst{1} \and Petia Radeva\inst{1,2}}
\authorrunning{Marc Bola\~nos et al.} 
%
\tocauthor{Marc Bola\~nos, Aina Ferr\`a and Petia Radeva}
\institute{Universitat de Barcelona, Barcelona, Spain
\and
Computer Vision Center, Bellaterra, Spain\\
\email{\{marc.bolanos, aferrama10.alumnes, petia.ivanova\}@ub.edu}}

\maketitle              

\vspace{-1em}
\begin{abstract}
   Automatically constructing a food diary that tracks the ingredients consumed can help people follow a healthy diet.
We tackle the problem of food ingredients recognition as a multi-label 
learning problem. We propose a method for adapting a highly performing state of the art CNN in order to act as a multi-label predictor for learning recipes in terms of their list of ingredients. 
We prove that our model is able to, given a picture, predict its list of ingredients, even if the recipe corresponding to the picture has never been seen by the model.    
We make public two new datasets suitable for this purpose. 
Furthermore, we prove that a model trained with a high variability of recipes and ingredients is able to generalize better on new data, and visualize how it specializes each of its neurons to different ingredients. 
\end{abstract}

\input{1_introduction.tex}
\input{2_related_work.tex}
\input{3_methodology.tex}
\input{4_results.tex}
\input{6_conclusions.tex}

{\small
\bibliographystyle{plain}
\bibliography{egbib}
}

\clearpage
\addtocmark[2]{Author Index} 
\renewcommand{\indexname}{Author Index}
\printindex
\clearpage
\addtocmark[2]{Subject Index} 
\markboth{Subject Index}{Subject Index}
\renewcommand{\indexname}{Subject Index}
\end{document}

%% file: 1_introduction.tex
\vspace{-2em}
\section{Introduction} \label{sec:introduction}


People's awareness about their nutrition habits is increasing 
either because they suffer from some kind of food intolerance; they have mild or severe weight problems; or they 
are simply interested in keeping a healthy diet. This increasing awareness is also being reflected in the technological world. Several applications exist for manually keeping track of what we eat, 
but they rarely offer any automatic mechanism for easing the tracking of the nutrition habits \cite{aizawa2015foodlog}. Tools for automatic food and ingredient recognition could heavily alleviate the problem. 

Since the reborn of Convolutional Neural Networks (CNNs), 
several works have been proposed to ease the creation of nutrition diaries.
The most widely spread approach is food recognition \cite{martinel2016wide}. 
These proposals allow to recognize the type of food present in an image and, consequently, could allow to approximately guess the ingredients contained and the overall nutritional composition. 
The main problem of these approaches is that no dataset covers the high amount of existent types of dishes worldwide (more than 8,000 according to Wikipedia).

On the other hand, a clear solution for this problem can be achieved if we formulate the task as an ingredients recognition problem instead \cite{chen2016deep}. Although tens of thousands of types of dishes exist, in fact they are composed of a much smaller  number of ingredients, which at the same time define the nutritional composition of the food. If we formulate the problem from the ingredients recognition perspective, we must consider the difficulty of distinguishing the presence of certain ingredients in cooked dishes.
Their visual appearance can greatly vary from one dish to another (e.g. the appearance of the ingredient 'apple' in an 'apple pie', an 'apple juice' or a 'fresh apple'), and in some cases they can even be invisible at sight without the proper knowledge of the true composition of the dish.
An additional benefit of approaching the problem from the ingredients recognition perspective is that, unlike in food recognition, it has the potential to predict valid outputs on data that has never been seen by the system. 

In this paper, we explore the problem of food ingredients recognition from a multi-label perspective by proposing a model based on CNNs that allows to discover the ingredients present in an image even if they are not visible to the naked eye. We present two new datasets for tackling the problem and prove that our method is capable of generalizing to new data that has never been seen by the system. Our contributions are four-fold.
  1) Propose a model for food ingredients recognition;
  2) Prove that by using a varied dataset of images and their associated ingredients, the generalization capabilities of the model on never seen data can be greatly boosted;
  3) Delve into the inner layers of the model for analysing the ingredients specialization of the neurons;
  and 4) Release two datasets for ingredients recognition.

This paper is organized as follows: in Section \ref{sec:related_work}, we review the state of the art; in Section \ref{sec:methodology}, explain our methodology; in Section \ref{sec:results}, we present our proposed datasets, show and analyse the results of the experiments performed, as well as interpret the predictions; 
and in Section \ref{sec:conclusions}, we draw some conclusions.

%% file: 2_related_work.tex
\vspace{-1em}
\section{Related work} 
\label{sec:related_work}
\vspace{-1em}

\vspace{.5em}
\textbf{Food analysis.} 
Several works have been published on applications related to automatic food analysis. Some of them proposed food detection models 
\cite{aguilar2017exploring} in order to distinguish when there is food present in a given image. Others focused on developing food recognition algorithms, either using conventional hand-crafted features, or powerful deep learning models \cite{martinel2016wide}. 
Others have applied food segmentation \cite{shimoda2015cnn}; 
use multi-modal data (i.e. images and recipe texts) for recipe recognition \cite{wang2015recipe}; tags from social networks for food characteristics perception \cite{ofli2017saki}; 
food localization and recognition in the wild for egocentric vision analysis \cite{bolanos2016simultaneous}, etc.

\vspace{.5em}
\textbf{Multi-Label learning.} Multi-label learning \cite{tsoumakas2006multi} consists in predicting more than one output category for each input sample. Thus, the problem of food ingredients recognition can be treated as a multi-label learning problem.
Several works 
\cite{wang2016cnn} argued that, when working with CNNs, they have to be reformulated for dealing with multi-label learning problems.
Some multi-label learning works have already been proposed for restaurant classification. 
So far, only one paper \cite{chen2016deep} has been proposed related to ingredients recognition. Their dataset, composed of 172 food types, was manually labelled considering visible ingredients only, which limits it to find 3 ingredients on average.
Furthermore, they propose a double-output model for simultaneous food type recognition and multi-label ingredients recognition. Although, 
the use of the food type for optimizing the model limits its capability of generalization only to seen recipes and food types. This fact becomes an important handicap in a real-world scenario when dealing with new recipes.
As we demonstrate in Sections \ref{subsec:results} and \ref{subsec:visualization}, unlike \cite{chen2016deep}, our model is able to: 1) recognize the ingredients appearing in unseen recipes (see Fig.\ref{fig:ingredients_recipes5k});  2) learn abstract representations of the ingredients directly from food appearance (see Fig.\ref{fig:neuron_activations}); and 3) infer invisible ingredients.

\vspace{.5em}
\textbf{Interpreting learning through visualization.} 
Applying visualization techniques is an important aspect in order to interpret what has been learned by our model. The authors in \cite{yosinski2015understanding}
have focused on proposing new ways of performing this visualization. At the same time, they have proven that CNNs have the ability to learn high level representations of the data and even hidden interrelated information, which can help us when dealing with ingredients that are apparently invisible in the image.

%% file: 3_methodology.tex
\vspace{-1em}
\section{Methodology} 
\label{sec:methodology}


\textbf{Deep multi-ingredients recognition.} Most of the top performing CNN architectures have been originally proposed and intended for the problem of object recognition. At the same time, they have been proven to be directly applicable to other related classification tasks and have served as powerful pre-trained models for achieving state of the art results. In our case, we compared either using the InceptionV3 \cite{szegedy2016rethinking} or the ResNet50 \cite{he2016deep} as the basic architectures for our model. We pre-trained it on the data from the ILSVRC challenge \cite{russakovsky2015imagenet} and modified the last layer for applying a multi-label classification over the $N$ possible output ingredients.
When dealing with classification problems, CNNs typically use the softmax activation in the last layer.
%
%
The softmax function allows to obtain a probability distribution for the input sample $x$ over all possible outputs and thus, predicts the most probable outcome, $\hat{y}_x = \argmax_{y_i} P(y_i|x)$.

The softmax activation is usually combined with the categorical cross-entropy loss function $L_c$ during model optimization, which penalizes the model when the optimal output value is far away from 1:
\begin{equation}
\vspace{-1em}
L_c = - \sum_x \log(P(\hat{y}_x|x)).
\end{equation}
In our model, we are dealing with ingredients recognition in a multi-label framework. Therefore, the model must predict for each sample $x$ a set of outputs represented as a binary vector $\hat{Y}_x = \{\hat{y}_x^1, ..., \hat{y}_x^N\}$, where $N$ is the number of output labels and each $\hat{y}_x^i$ is either 1 or 0 depending if it is present or not in sample $x$. For this reason, instead of softmax, we use a sigmoid activation function:
\begin{equation}
P(y_i|x) = \frac{1}{1-\exp^{-f(x)_i}}
\end{equation}
which allows to have multiple highly activated outputs. 
For considering the binary representation of $\hat{Y}_x$, we chose the binary cross-entropy function $L_b$ \cite{buja2005loss}:
\begin{equation}
\vspace{-1em}
L_b = - \sum_x \sum_{i}^N (\hat{y}_x^i\cdot log(P(y_i|x)) + (1 - \hat{y}_x^i) \cdot log(1 - P(y_i|x)))
\end{equation}
which during backpropagation rewards the model when the output values are close to the target vector $\hat{Y}_x$ (i.e. either close to 1 for positive labels or close to 0 for negative labels).

%% file: 4_results.tex
\vspace{-1em}
\section{Results} 
\label{sec:results}
\vspace{-1em}


In this section, we describe the two datasets proposed for the problem of food ingredients recognition. Later we describe our experimental setup 
and at the end, we present the final results obtained both for ingredients recognition on known classes as well as recognition results for 
generalization on samples never seen by the model.


\vspace{-1em}
\subsection{Datasets}

In this section we describe the datasets proposed for food ingredients recognition and the already public datasets used.

\vspace{.5em}
\textbf{Food101 \cite{bossard2014food}}
is one of the most widely extended datasets for food recognition. It consists of 101,000 images equally divided in 101 food types.

\vspace{.5em}
\textbf{Ingredients101}\footnote{\url{http://www.ub.edu/cvub/ingredients101/}} is a dataset for ingredients recognition that we constructed and make public in this article. It consists of the list of most common ingredients for each of the 101 types of food contained in the Food101 dataset, making a total of 446 unique ingredients (9 per recipe on average). The dataset was divided in training, validation and test splits making sure that the 101 food types were balanced. We make public the lists of ingredients together with the train/val/test split applied to the images from the Food101 dataset.

\vspace{.5em}
\textbf{Recipes5k}\footnote{\url{http://www.ub.edu/cvub/recipes5k/}} is a dataset for ingredients recognition with 4,826 unique recipes composed of an image and the corresponding list of ingredients. It contains a total of 3,213 unique ingredients (10 per recipe on average). 
Each recipe is an alternative way to prepare one of the 101 food types in Food101. Hence, it captures at the same time the intra-class variability and inter-class similarity of cooking recipes. The nearly 50 alternative recipes belonging to each of the 101 classes were divided in train, val and test splits in a balanced way. We make also public this dataset together with the splits division.
A problem when dealing with the 3,213 raw ingredients is that many of them are sub-classes (e.g. 'sliced tomato' or 'tomato sauce') of more general versions of themselves (e.g. 'tomato'). Thus, we propose a simplified version by applying a simple removal of overly-descriptive particles\footnote{\url{https://github.com/altosaar/food2vec}} (e.g. 'sliced' or 'sauce'), resulting in 1,013 ingredients used for additional evaluation (see Section \ref{subsec:results}). 



We must note the difference between our proposed datasets and the one from \cite{chen2016deep}. While we consider any present ingredient in a recipe either visible or not, the work in \cite{chen2016deep} only labelled manually the visible ingredients in certain foods. Hence, a comparison between both works is infeasible.

\begin{figure}[!ht]
\vspace{-2em}
\centering	
  \begin{subfigure}{.48\textwidth}
  	\centering	
      \includegraphics[trim={9cm 0.5cm 0 0.2cm},clip,width=0.8\linewidth]{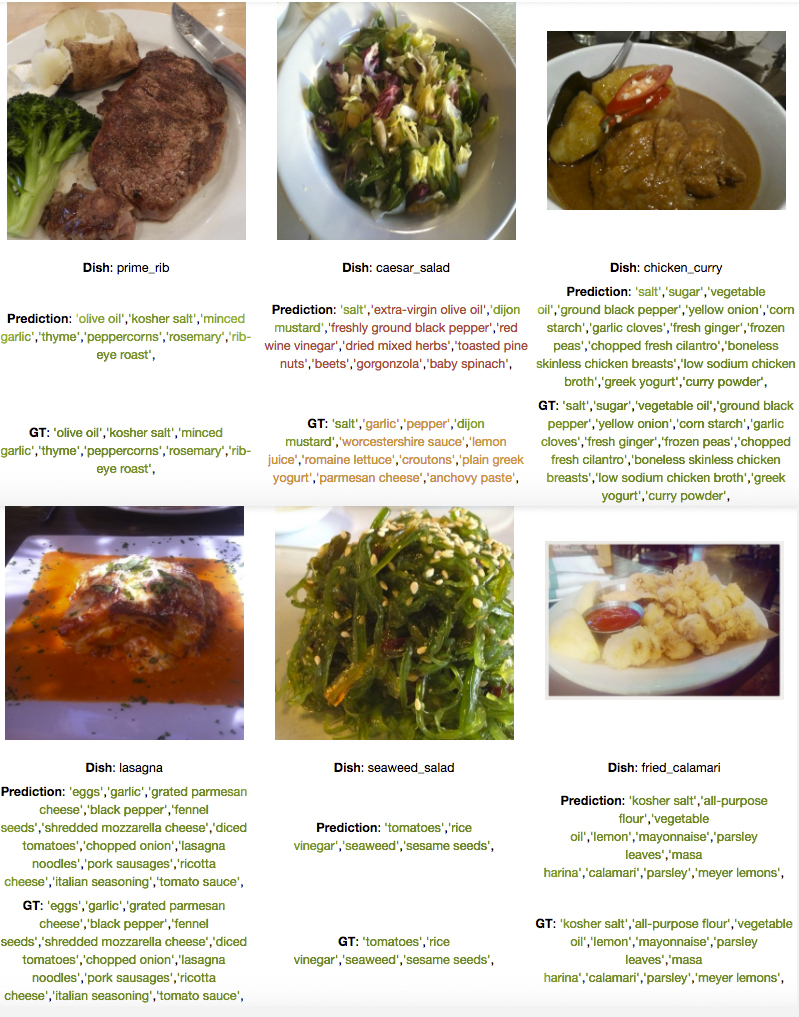}	\caption{\label{fig:ingredients_101_results} Ingredients101 samples.}
  \end{subfigure}\hfill%
  \begin{subfigure}{.48\textwidth}
  	\centering	
      \includegraphics[trim={0 23cm 0 0},clip,width=0.9\linewidth]{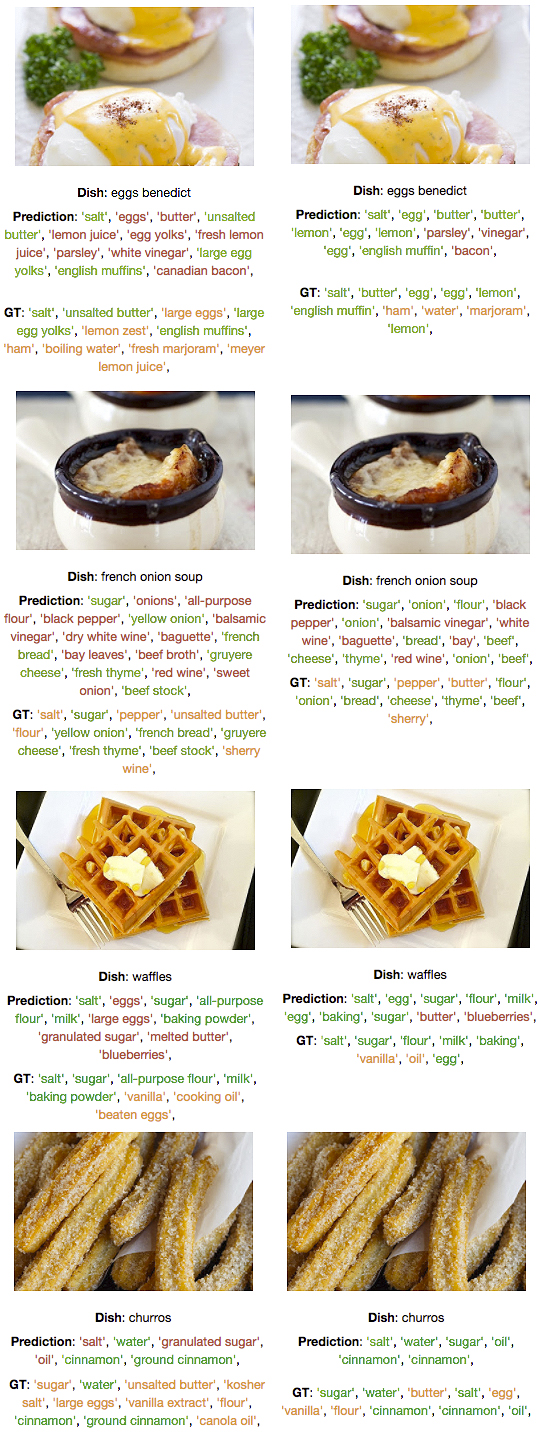}
      \caption{\label{fig:ingredients_recipes5k} Recipes5k using the fine-grained 3,213 ingredients (left), and using the 1,013 simplified ingredients (right).}
    \end{subfigure}
    \caption{Our method's results. TPs in green, FPs in red and FNs in orange.}
    \vspace{-1em}
\end{figure}

\begin{table*}[ht]
\begin{center}
\begin{tabular}{l c c c c c c}
 & \multicolumn{3}{c}{Validation} & \multicolumn{3}{c}{Test} \\ \cmidrule(lr){2-4} \cmidrule(lr){5-7}
 & Prec & Rec & $F_1$ & Prec & Rec & $F_1$ \\ 
\hline
Random prediction & 2.05 & 2.01 & 2.03 & 2.06 & 2.01 & 2.04 \\
InceptionV3 + Ingredients101 & 80.86 & 72.12 & 76.24 & 83.51 & \textbf{76.87} & 80.06\\
ResNet50 + Ingredients101& 84.80 & 67.62 & 75.24 & \textbf{88.11} & 73.45 & \textbf{80.11}\\
\hline
\end{tabular}
\end{center}
\caption{\label{tab:results_ingredients101}Ingredients recognition results obtained on the dataset Ingredients101. Prec stands for \textit{Precision}, Rec for \textit{Recall} and $F_1$ for \textit{$F_1$ score}. All measures reported in \%. The best test results are highlighted in boldface.}
\vspace{-1em}
\end{table*}

\begin{table*}[ht]
\vspace{-1em}
\begin{center}
\begin{tabular}{l c c c c c c}
 & \multicolumn{3}{c}{Validation} & \multicolumn{3}{c}{Test} \\ \cmidrule(lr){2-4} \cmidrule(lr){5-7}
 & Prec & Rec & $F_1$ & Prec & Rec & $F_1$ \\ 
\hline
Random prediction & 0.33 & 0.32 & 0.33 & 0.54 & 0.53 & 0.53 \\
InceptionV3 + Ingredients101 & & & & 23.80 & 18.24 & 20.66 \\
ResNet50 + Ingredients101 & & & & 26.28 & 16.85 & 20.54 \\
InceptionV3 + Recipes5k & 36.18 & 20.69 & 26.32 & 35.47 & \textbf{21.00} & \textbf{26.38}\\
ResNet50 + Recipes5k & 38.41 & 19.67 & 26.02 & \textbf{38.93} & 19.57 & 26.05 \\
\hline
Random prediction & 6.27 & 6.29 & 6.28 & 6.14 & 6.24 & 6.19 \\
InceptionV3 + Ingredients101 & & & & 44.01 & 34.04 & 38.39 \\
ResNet50 + Ingredients101 & & & & 47.53 & 30.91 & 37.46 \\
InceptionV3 + Recipes5k & 56.77 & 31.40 & 40.44 & 55.37 & 31.52 & 40.18\\
ResNet50 + Recipes5k & 56.73 & 28.07 & 37.56 & \textbf{58.55} & 28.49 & 38.33 \\
InceptionV3 + Recipes5k simplified & 53.91 & 42.13 & 47.30 & 53.43 & \textbf{42.77} & \textbf{47.51}\\
\hline
\end{tabular}
\end{center}
\caption{\label{tab:results_recipes5k}Ingredients recognition results on Recipes5k (top) and on Recipes5k simplified (bottom). Prec stands for \textit{Precision}, Rec for \textit{Recall} and $F_1$ for \textit{$F_1$ score}. All measures reported in \%. Best test results are highlighted in boldface.}
\vspace{-2.5em}
\end{table*}


\begin{figure*}[!ht]
	\centering	\includegraphics[trim={0 18cm 0 0},clip,width=0.65\textwidth]{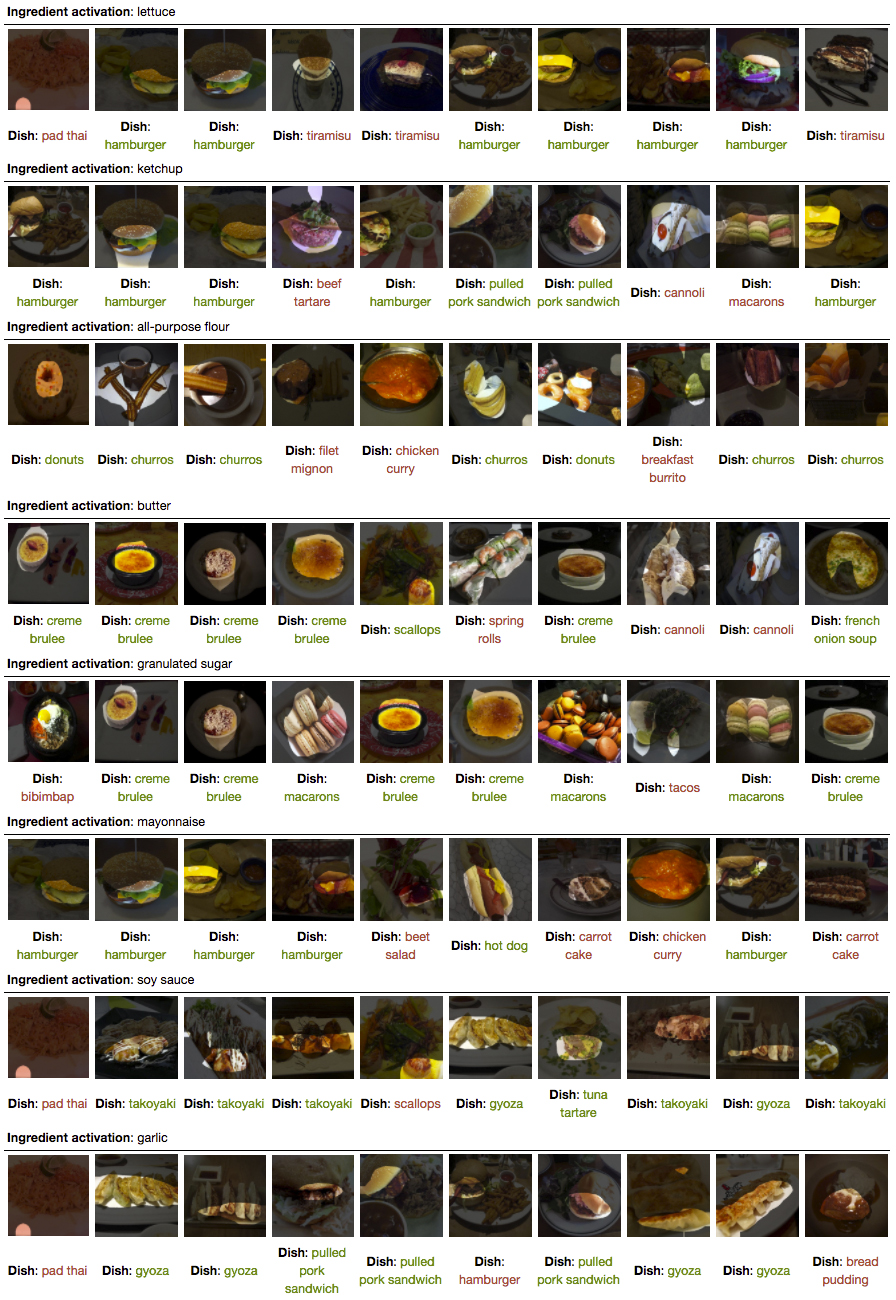}	\caption{\label{fig:neuron_activations} Visualization of neuron activations. Each row is associated to a specific neuron from the network. The images with top activation are shown as well as the top ingredient activation they have in common. The name of their respective food class is only for visualization purposes and is displayed in green if the recipe contains the top ingredient. Otherwise, it is shown in red.}
    \vspace{-2em}
\end{figure*}


\vspace{-1em}
\subsection{Experimental setup}

Our model was implemented in Keras\footnote{\url{www.keras.io}}, using 
Theano 
as backend. 
Next, we detail the different configurations and tests performed.
\textbf{Random prediction}: (baseline) a set of $K$ labels are generated uniformly distributed among all possible outputs. $K$ depends on the average number of labels per recipe in the corresponding dataset.
\textbf{InceptionV3 + Ingredients101}: InceptionV3 model pre-trained on ImageNet and adapted for multi-label learning. 
\textbf{ResNet50 + Ingredients101}: ResNet50 model pre-trained on ImageNet and adapted for multi-label learning. 
\textbf{InceptionV3 + Recipes5k}: InceptionV3 model pre-trained on InceptionV3 + Ingredients101. 
\textbf{ResNet50 + Recipes5k}: ResNet50 model pre-trained on ResNet50 + Ingredients101. 

\vspace{-1.5em}
\subsection{Experimental results}\label{subsec:results}
\vspace{-0.5em}

In Table \ref{tab:results_ingredients101}, we show the ingredient recognition results on the Ingredients101 dataset. In Fig.\ref{fig:ingredients_101_results} some qualitative results are shown. Both the numerical results and the qualitative examples prove the high performance of the models in most of the cases. 
Note that although a multi-label classification is being applied, considering that all the samples from a food class share the same set of ingredients, the model is indirectly learning the inherent food classes. Furthermore, looking at the results on the Recipes5k dataset in Table \ref{tab:results_recipes5k} (top), we can see that the very same model obtains reasonable results even considering that it was not specifically trained on that dataset. Note that only test results are reported for the models trained on Ingredients101 because we only intend to show its generalization capabilities on new data.

Comparing the results with the models specifically trained on Recipes5k, it appears that, as expected, a model trained on a set of samples with high variability of output labels 
is more capable of obtaining high 
results on never seen recipes. Thus, it is more capable of generalizing on unseen data. 

Table \ref{tab:results_recipes5k} (bottom) shows the results on the Recipes5k dataset with a simplified list of ingredients. Note that for all tests, the list was simplified only during the evaluation procedure for maintaining the fine-grained recognition capabilities of the model, with the exception of \textit{Inception V3 + Recipes5k simplified}, where the simplified set was also used for training. The simplification of the ingredients list enhances the capabilities of the model when comparing the results, reaching more than 40\% in the $F_1$ metric and 47.5\% also training with them. 

Fig.\ref{fig:ingredients_recipes5k} shows a comparison of the output of the model either using the  fine-grained or the simplified list of ingredients.
Overall, although usually only a single type of semantically related fine-grained ingredients (e.g. 'large eggs', 'beaten eggs' or 'eggs') appears at the same time in the ground truth, it seems that the model is inherently learning an embedding of the ingredients. Therefore, it is able to understand that some fine-grained ingredients are related and predicts them at once in the fine-grained version (see waffles example).

\vspace{-1em}
\subsection{Neuron representation of ingredients}\label{subsec:visualization}
\vspace{-0.5em}

When training a CNN model, it is important to understand what it is able to learn and interpret from the data. To this purpose, we visualized the activations of certain neurons of the network in order to interpret what is it able to learn.

Fig.\ref{fig:neuron_activations} shows the results of this visualization. As we can see, it appears that certain neurons of the network are specialized to distinguish specific ingredients. For example, most images of the 1st and 2nd rows illustrate that the characteristic shape of a hamburger implies that it will probably contain the ingredients 'lettuce' and 'ketchup'. Also, looking at the 'granulated sugar' row, we can see that the model learns to interpret the characteristic shape of \textit{creme brulee} and \textit{macarons} as containing sugar, although it is not specifically seen in the image.

%% file: 6_conclusions.tex
\vspace{-1em}
\section{Conclusions and future work} \label{sec:conclusions}
\vspace{-0.5em}

Analysing both the quantitative and qualitative results, we can conclude that the proposed model and the two datasets published offer very promising results for the multi-label problem of food ingredients recognition. 
Our proposal allows to obtain 
great generalization results on unseen recipes and sets the basis for applying further, more detailed food analysis methods.
As future work, we will create a hierarchical structure \cite{wu2016learning} relationship of the existent ingredients and extend the model to utilize this information.

\vspace{-1em}